\documentclass[runningheads]{llncs}
\usepackage{graphicx}
\usepackage{amsmath}
\newcommand{\superscript}[1]{\ensuremath{^\textrm{#1}}}

\newcommand{\ie}{i.\,e.~}

\begin{document}
\title{Localizing dexterous surgical tools in X-ray for image-based navigation\thanks{This work was supported in part by NIH R01 EB023939 and Johns Hopkins University internal funding sources.}}
\titlerunning{Localizing dexterous surgical tools in intra-operative X-ray}
%
\author{Cong~Gao\superscript{1,2,*} \and
Mathias~Unberath\superscript{1,2,*} \and
Russell~Taylor\superscript{1,2} \and Mehran~Armand\superscript{2,3,4}}
\authorrunning{Gao, Unberath, Taylor, and Armand}
%
\institute{
\superscript{1} Department of Computer Science, Johns Hopkins University
\\
\superscript{2} Laboratory for Computational Sensing and Robotics, Johns Hopkins University
\\
\superscript{3} Johns Hopkins University Applied Physics Laboratory
\\
\superscript{4} Department of Orthopaedic Surgery, Johns Hopkins Hospital
\\
\superscript{*} Joint first authors.
\\
\email{cgao11@jhu.edu}
}%
\maketitle              

\section{Purpose}
\label{Purpose}
Continuum dexterous manipulators (CDMs), commonly referred to as snake-like robots, have demonstrated great premise for minimmally-invasive procedures~\cite{walker2016snake,burgner2015continuum}. Recent innovations have made CDMs appropriate for use in orthopedic surgery~\cite{kutzer2011design,murphy2014design,alambeigi2017curved}. One key challenge of using CDMs is performing precise intra-operative control guided by a pre-operative patient-specific plan, conceived based on 3D imaging and potentially bio-mechanical analysis. To this end, the calibration loop of robot base to end-effector to patient anatomy must be closed, and an accurate kinematic deformation estimation of the CDM is required.\\
X-ray image based surgical tool navigation has received increasing interest since it is fast and supplies accurate images of deep seated structures. Typically, recovering the 6 degree of freedom (DOF) rigid pose and deformation of tools with respect to the X-ray camera can be accurately achieved through intensity-based 2D/3D registration of 3D images or models to 2D X-rays~\cite{markelj2012review}. However, it is well known that the capture range of image-based 2D/3D registration is inconveniently small suggesting that automatic and robust initialization strategies are of critical importance. 
Consequently, this manuscript describes a first step towards leveraging semantic information of the imaged object to initialize 2D/3D registration within the capture range of image-based registration by performing concurrent segmentation and localization of the CDM in X-ray images. 

\section{Methods}
\begin{figure}[tb]
 \centering
    \includegraphics[width=0.7\textwidth]{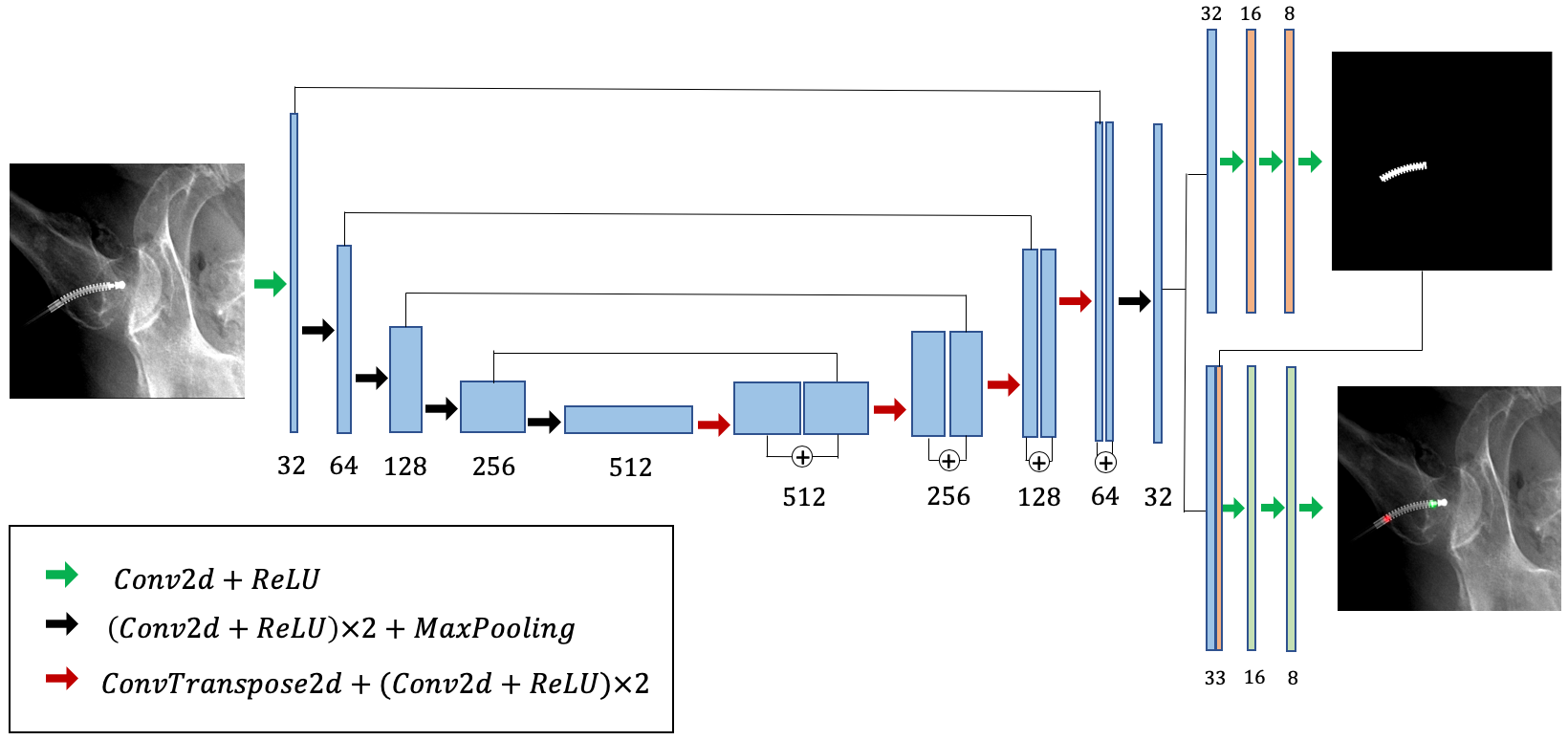}
    \caption{Network architecture used for concurrent segmentation and landmark detection.}
    \label{fig:network}
\end{figure}

We seek to train a convolutional neural network (ConvNet) to localize CDMs in X-ray images. The CDM developed by our group considered here is fabricated using Nitinol. Its outer diameter is $6$\,mm and it includes an instrument channel with a diameter of $4$\,mm. A set of 26 alternating notches on the body of the CDM allow for single-plane bending~\cite{kutzer2011design,murphy2014design}. 
Due to the unavailability of annotated X-ray images to train ConvNets, we rely on DeepDRR~\cite{unberath2018deepdrr}, a framework for physics-based rendering of digitally reconstructed radiographs (DRRs, \ie synthetic fluoroscopic images) from 3D CT. DeepDRR accurately accounts for energy- and material-dependence of X-ray attenuation, scattering, and noise. Recent work~\cite{unberath2018deepdrr,bier2018landmarks} demonstrated that ConvNets trained on DeepDRRs generalize to clinically acquired X-ray images without re-training, motivating its use for the application proposed here. The simulation of the CDM mainly consists of two parts: 1) body and base of the CDM plus an extended shaft; and 2) inserted tool and drill. Following previous work on kinematic modeling of this CDM~\cite{otake2014piecewise}, we assume that the joint angle changes smoothly from one joint to the next. Angles are parameterized as cubic spline of $n =5$ equally distributed control points, $\tau_i$, along the central axis of the CDM. The rigid pose of the CDM relative to X-ray camera is represented by translation and rotation in $x$, $y$, and $z$ axes, defining the total parameter space as $\theta=\{t_x, t_y, t_z, r_x, r_y, r_z, c\},\ (c=\{\tau_1,...,\tau_n\})$.\\
Given a 3D CT of the lower torso, we manually define a rigid transformation such that the CDM model is enclosed in the femur, simulating applications in core decompression and fracture repair~\cite{alambeigi2017curved,alambeigi2018inroads}. DeepDRR uses voxel representation, so the CDM surface model is voxelized with high resolution to preserve details of the notches. At positions where the CT volume exhibits overlap with the CDM, CT values are omitted to model drilling. From the above volumes and coordinate transforms, we use DeepDRR to generate 1) realistic X-ray images, 2) 2D segmentation masks of the CDM end-effector, and 3) 2D locations of two key landmarks. Our segmentation target region covers the 26 alternating notches which discerns the CDM from other surgical tools. The two landmarks are defined as 1) the middle of the 2 conjunction points between the first notch and the base and 2) the center of the distal plane of the last notch, \ie start and end point of CDM centerline. 
The simulated X-rays have $512\times512$ pixels with an isotropic pixel size of $(0.62$\,mm$)^2$. DRRs are converted to line-integral domain to decrease the dynamic range and then normalized to $[-1,1]$. Landmark coordinates are transformed to belief maps expressed as Gaussian distributions ($\sigma=5$\,pixels) around the true location. 
Data generation was done as follows: A total of 5 lower-limbs CTs ($512\times512\times2590$\,voxel, $0.85$~mm$^3$/voxel) were included in the experiment and centered around the pelvis. The CDM volume was manually aligned with the left/right femur to mimic our clinical usecase. Then, CDM shapes and rigid X-ray source and volume poses were sampled randomly: Source-to-detector distance was fixed to $1200$\,mm while source-to-isocenter distance was $\in[400\,\text{mm}, 500\,\text{mm}]$. Source rotation in LAO/RAO was $\in[0^\circ, 360^\circ]$ and in CRAN/CUAD $\in[75^\circ, 105^\circ]$. Volume translation was $\in[-20,20]$\,mm in all axes. CDM shapes were defined by randomly sampling control point angles $\in[-7.9^\circ, 7.9^\circ]$. We sampled a total of $1,000$ random configurations per femoral head ($10,000$ total) to render synthetic images. CTs were split $4:1$ into training:testing, and within the training dataset $10:1$ into training:validation. We also manually annotated 87 X-ray images of a real CDM drilling in femoral bone specimens for quantitative evaluation on cadaveric data.\\
Inspired by the work of~\cite{laina2017concurrent}, we design a ConvNet-based auto-encoder like architecture with skip connections, and split the connection from the last feature layer to perform two tasks concurrently, \ie segmentation and landmark detection. Fig.~\ref{fig:network} illustrates the ConvNet architecture used here. In the decoder, we repeat the connection of 2D convolutional layer and maxpooling layer four times to abstract a feature representation with $512$ channels. In the decoder part, we concatenate the upsampled features and features from the same level in the encoding stage. The final decoded $32$ channel feature layer is shared across the segmentation path and the localization path. The final output of the segmentation mask is backward concatenated with this shared feature to boost the localization task. We chose Dice loss to train the segmentation task and the standard $L_2$ loss for the localization task. Learning rate was initialized with $0.001$ and decayed by $0.1$ every $10$ epochs.      
 
\section{Results}
\begin{figure}[tb]
 \centering
    \includegraphics[width=\textwidth]{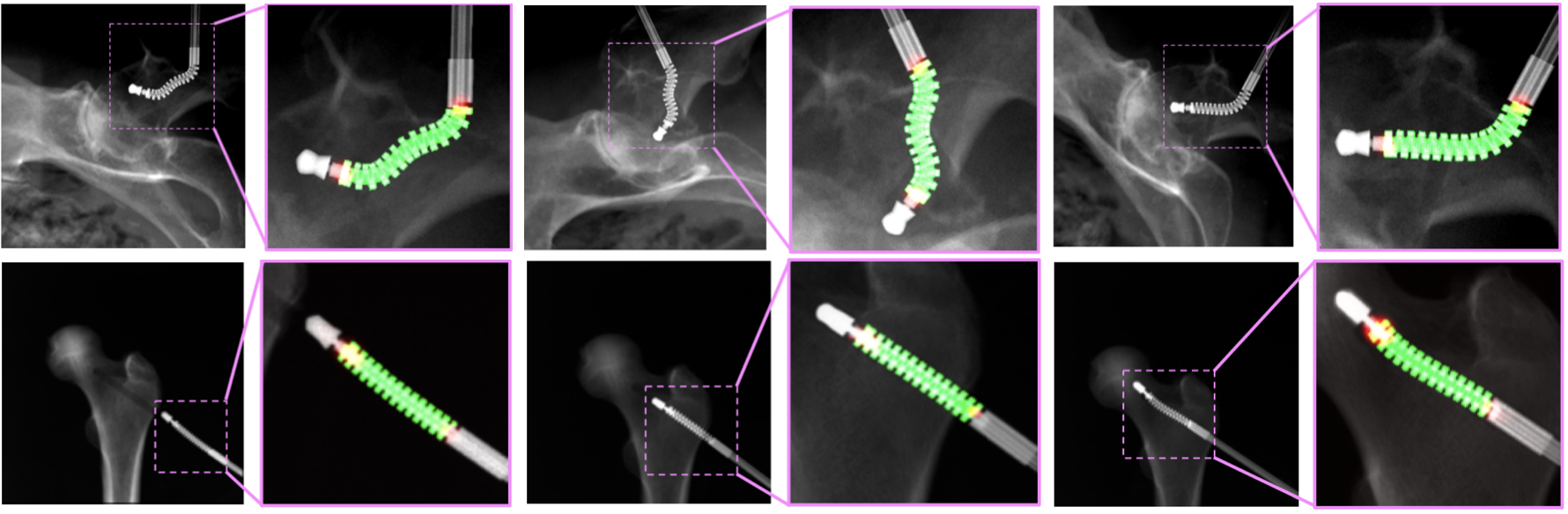}
    \caption{Representative examples of segmentation and landmark detection performance on synthetic (upper row) and real \emph{ex vivo} data (lower row). The predicted segmentation and landmarks are shown as green and red overlay, respectively.}
    \label{fig:detection_result}
\end{figure}

The segmentation accuracy is computed as the Dice score of mask prediction. Landmark detection accuracy is reported as the $L_2$ distance in millimeters. We first evaluated the network on the synthetic dataset where exact groundtruth was known. The mean Dice score was $0.996\pm0.001$ and the mean $L_2 $ distance was $0.365\pm0.345$\,mm. On the manually annotated 87 \textit{ex vivo} X-ray images, the network achieved a mean Dice score of $0.915\pm0.063$ and mean $L_2 $ distance of $2.54\pm0.95$\,mm. The cadaveric data contained configurations never seen during training (\ie tool completely outside bone) that induced poor performance of our network, as reflected in the high standard deviations for the cadaveric dataset. Representative results are shown in Fig.~\ref{fig:detection_result}.

\section{Conclusions}
We presented a learning-based strategy to simultaneously localize and segment dexterous surgical tools in X-ray images. Our results on synthetic and \emph{ex vivo} data are promising and encourage training of our ConvNet on a more exhaustive dataset. We currently investigate how these results translate to other real data and investigate methods to use semantic information extracted by the proposed network to reliably and robustly initialize image-based 2D/3D registration. 
%
%
%
\bibliographystyle{splncs04}
%
\bibliography{bib_short}
\end{document}